\documentclass[letterpaper, 10 pt, conference]{ieeeconf}  

\IEEEoverridecommandlockouts                              

\usepackage{graphicx}
\usepackage{microtype}

\usepackage{pgfplots}
\usepackage{tikz}
\usetikzlibrary{calc}
\usetikzlibrary{positioning}
\usepgfplotslibrary{colorbrewer}

\usetikzlibrary{external}
\usepackage{amsmath}

\usepackage{subcaption}
\DeclareCaptionLabelSeparator{periodspace}{.\quad}
\usepackage{algorithm, algorithmic}
\floatstyle{plaintop}
\restylefloat{algorithm}

\usepackage{placeins}

\usepackage{amsmath}
\DeclareMathOperator*{\argmax}{arg\,max}

\title{\LARGE \bf Improving Sonar Image Patch Matching via Deep Learning}

\author{Matias Valdenegro-Toro$^{1}$
    \thanks{$^{1}$Matias Valdenegro-Toro is with Ocean Systems Laboratory,
        School of Engineering \& Physical Sciences, Heriot-Watt University, EH14 4AS, Edinburgh, UK
        {\tt\small m.valdenegro@hw.ac.uk}}%
}

\begin{document}

\maketitle
\thispagestyle{empty}
\pagestyle{empty}

\begin{abstract}
Matching sonar images with high accuracy has been a problem for a long time, as sonar images are inherently hard to model due to reflections, noise and viewpoint dependence. Autonomous Underwater Vehicles require good sonar image matching capabilities for tasks such as tracking, simultaneous localization and mapping (SLAM) and some cases of object detection/recognition. We propose the use of Convolutional Neural Networks (CNN) to learn a matching function that can be trained from labeled sonar data, after pre-processing to generate matching and non-matching pairs. In a dataset of 39K training pairs, we obtain 0.91 Area under the ROC Curve (AUC) for a CNN that outputs a binary classification matching decision, and 0.89 AUC for another CNN that outputs a matching score. In comparison, classical keypoint matching methods like SIFT, SURF, ORB and AKAZE obtain AUC 0.61 to 0.68. Alternative learning methods obtain similar results, with a Random Forest Classifier obtaining AUC 0.79, and a Support Vector Machine resulting in AUC 0.66.
\end{abstract}

\section{Introduction}

One of the basic problems in robotics is data association, where sensor readings have to be associated with previous measurements, as the combination of sensor data reduces noise and improves robot understanding of the world. Autonomous Underwater Vehicles (AUVs) constantly struggle with data association, as the underwater environment is very hostile for sensing. Some common robot tasks that require data association are tracking and simultaneous localization and mapping (SLAM). Object detection/recognition can also benefit from data association in the form of matching images if the task is to locate an object and only a single training sample is available.

Acoustic sensing (Sonar) is used in underwater environments as sound can travel large distances on water with little attenuation. Optical cameras are not an option as light is attenuated and absorbed by particles in the water column. Interpretation of acoustic images is not trivial as unwanted reflections, noise, and low signal-to-noise ratio (SNR) degrades the amount of information that the AUV can gather.

For sonar, matching image patches to known objects or landmarks in the environment is an important problem. Matching can also be formulated for other tasks such as mosaicing \cite{hurtos2012fourier}, where sonar images must be registered before being combined to improve SNR ratio and image resolution. Matching sonar images is difficult due to viewpoint dependence.

In this work, we propose the use of Convolutional Neural Networks (CNN) to learn a matcher for sonar images. Our objective is to produce a function that takes two sonar image patches and makes a binary decision: both images correspond to different views of the same object, or not. Matching should be possible even as insonification\footnote{Amount of acoustic signal that "illuminates" the target area by the sonar sensor.} varies due to AUV or sensor movement, different views, and object rotation or translation.

CNNs have obtained very good results \cite{lecun1998gradient} in different tasks that use optical color images, such as object recognition \cite{krizhevsky2012imagenet} and transfer learning \cite{razavian2014cnn}. We have previously evaluated CNNs for object recognition in sonar images and found that they also improve the state of the art \cite{valdenegro2016object}. CNNs have also been used to match patches from color images \cite{zagoruyko2015learning} with high accuracy. These results motivate us to use CNNs for sonar data, as the trained network can learn sonar-specific information directly from the data.

We show that we can build and train a CNN that matches Forward-Looking Sonar (FLS) image patches with high accuracy (AUC\footnote{Area under the ROC Curve} 0.91), surpassing the state of the art keypoint matchers such as SIFT and SURF (with AUC in the range 0.61 - 0.68).

Our contributions are: we propose an algorithm to generate matching pairs from labeled objects for training, we learn the matching function directly from labeled data, without any manual feature engineering, we show that it is possible to match sonar images with relatively high accuracy.

\section{Related Work}

Matching sonar images with high accuracy has been a unsolved problem for a long time \cite{negahdaripour2011dynamic} \cite{hurtos2012fourier} \cite{vandrish2011side}. This is due to specifics issues in sonar imaging, such as viewpoint dependence, non-uniform insonification, low signal-to-noise ratio, low resolution, and low feature repeatability \cite{hurtos2013automatic}. Most methods that are used for different kinds of matching in sonar imagery are not specifically designed for sonar (originally developed for optical images), and do not consider sonar-specific information.

Kim et al. \cite{kim2005mosaicing} matches keypoints detected with the Harris corner detection to register general sonar images. Vandrish et al. \cite{vandrish2011side} compares the use of SIFT \cite{lowe2004distinctive} with different feature methods for sidescan sonar image registration, concluding that for this task SIFT performs best. Hurtos et al. \cite{hurtos2012fourier} uses Fourier-based features for registration of Forward-Looking Sonar images, with great success. These kind of features could be used to make a matching decision, but they only work appropriately when rotation/translation between frames are small. Pham et al. \cite{pham2013guided} uses block-matching guided by a segmented sonar image with a Self-Organizing Map for registration and mosaicing of sidescan sonar images.

A large portion of the research about matching sonar images is devoted to registration and mosaicing \cite{kim2005mosaicing} \cite{hurtos2012fourier}. Both processes require many assumptions on the kind of images and their content, specially when considering non-uniform insonification and simple transformations between images.

In comparison, CNNs \cite{krizhevsky2012imagenet} have been used to compare and match color image patches. Zagoruyko et al. \cite{zagoruyko2015learning} uses CNNs trained on a dataset of 500K matched patches with high accuracy in tasks such as stereo matching and descriptor evaluation. Raw matching performance is also good, but only possible due to the availability of large labeled datasets.

Zbontar and LeCun \cite{zbontar2016stereo} also use CNNs for stereo matching, improving over the state of the art in several datasets. These recent results using CNNs motivate us to explore such algorithm for matching sonar images. CNNs have several advantages when applied to sonar imaging: they can learn sonar-specific information directly from raw data, they do not require feature engineering or specific data preprocessing, and they make little assumptions on input data.

\section{Matching Sonar Image Patches with CNNs}

\subsection{Training Data Generation}

Given a dataset containing labeled bounding boxes (including object classes), we generate matching and non-matching image pairs that are sampled from the dataset. We do this by using object class information to generate matching image pairs, and we also produce non-matching pairs that contain objects versus background. The dataset that we used for this purpose was originally designed for object detection. We generate the following kinds of pairs:

\begin{itemize}
    \item \textbf{Object vs Object, Same class}. A matching pair is generated from two objects of the same class. We sample two random image crops of objects in the same class and generate one pair, typically both crops corresponds to different perspectives of the same object, or different insonification levels from the sensor.
    \item \textbf{Object vs Object, Different class}. A non-matching pair is generated from two objects from different classes. This makes the assumption that objects in the dataset are not similar across different classes.
    \item \textbf{Object vs Background}. A non-matching pair is generated by sampling one background patch that has IoU score lower than 0.1 with the ground truth and generating a pair with a random object image crop.
\end{itemize}

\begin{figure}
    \centering
    \begin{subfigure}[b]{0.5 \textwidth}
        \includegraphics[width=0.15 \textwidth]{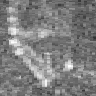}
        \includegraphics[width=0.15 \textwidth]{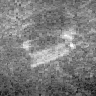} \;
        \includegraphics[width=0.15 \textwidth]{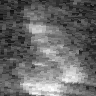}
        \includegraphics[width=0.15 \textwidth]{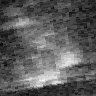} \;
        \includegraphics[width=0.15 \textwidth]{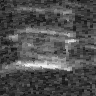}
        \includegraphics[width=0.15 \textwidth]{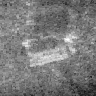} \\
        
        \includegraphics[width=0.15 \textwidth]{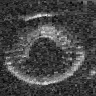}
        \includegraphics[width=0.15 \textwidth]{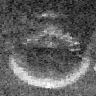} \;
        \includegraphics[width=0.15 \textwidth]{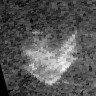}
        \includegraphics[width=0.15 \textwidth]{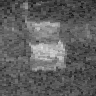} \;
        \includegraphics[width=0.15 \textwidth]{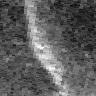}
        \includegraphics[width=0.15 \textwidth]{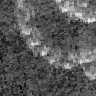}
        
        \caption{Object-Object Matches}
    \end{subfigure}        
    \begin{subfigure}[b]{0.5 \textwidth}
        \includegraphics[width=0.15 \textwidth]{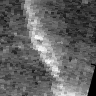}
        \includegraphics[width=0.15 \textwidth]{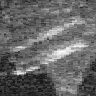} \;
        \includegraphics[width=0.15 \textwidth]{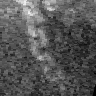}
        \includegraphics[width=0.15 \textwidth]{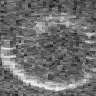} \;
        \includegraphics[width=0.15 \textwidth]{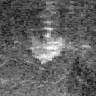}
        \includegraphics[width=0.15 \textwidth]{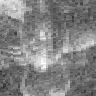} \\
        
        \includegraphics[width=0.15 \textwidth]{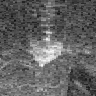}
        \includegraphics[width=0.15 \textwidth]{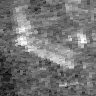} \;
        \includegraphics[width=0.15 \textwidth]{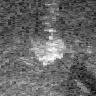}
        \includegraphics[width=0.15 \textwidth]{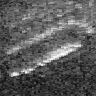} \;
        \includegraphics[width=0.15 \textwidth]{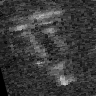}
        \includegraphics[width=0.15 \textwidth]{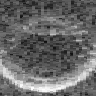}
        
        \caption{Object-Object Non-Matches}
    \end{subfigure}           
    \begin{subfigure}[b]{0.5 \textwidth}
        \includegraphics[width=0.15 \textwidth]{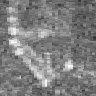}
        \includegraphics[width=0.15 \textwidth]{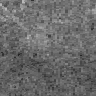} \;
        \includegraphics[width=0.15 \textwidth]{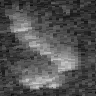}
        \includegraphics[width=0.15 \textwidth]{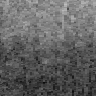} \;
        \includegraphics[width=0.15 \textwidth]{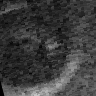}
        \includegraphics[width=0.15 \textwidth]{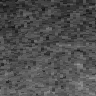} \\
        
        \includegraphics[width=0.15 \textwidth]{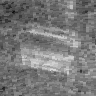}
        \includegraphics[width=0.15 \textwidth]{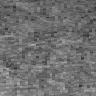} \;
        \includegraphics[width=0.15 \textwidth]{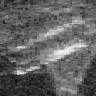}
        \includegraphics[width=0.15 \textwidth]{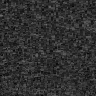} \;
        \includegraphics[width=0.15 \textwidth]{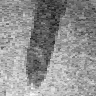}
        \includegraphics[width=0.15 \textwidth]{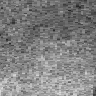}
        
        \caption{Object-Background Non-Matches}
    \end{subfigure}           
    
    \caption{A small sample sonar image patches labeled as matching or non-matching that were generated by our algorithm. These patches were captured with an ARIS Explorer 3000 Forward-Looking Sonar.}
    \label{flsPatchesSample}
\end{figure}

As the number of possible non-matching pairs is very large, we balance matches (positive) and non-matches (negative) samples to be $1:1$. This is done by sampling 10 matches per object, 5 non-matches between objects of different class, and 5 non-matches with background. The detailed algorithm is presented in Algorithm \ref{pairGenerationAlgorithm}. We generate pairs of $96 \times 96$ image crops, as this is the most appropriate size for the objects in our dataset. A small sample of generated pairs is shown in Fig. \ref{flsPatchesSample}.

\begin{algorithm}
    \caption{Training Data Generation}
    \label{pairGenerationAlgorithm}
    
    \begin{algorithmic}[1]
        \renewcommand{\algorithmicrequire}{\textbf{Input:}}
        \renewcommand{\algorithmicensure}{\textbf{Output:}}
        
        \REQUIRE Labeled Image dataset $I$ with bounding boxes $B_i$ and class labels $C_i$, number of positive samples $S_p$, number of negative samples $S_n$.
        
        \ENSURE List of matching pairs $L_m$ and list of non-matching pairs $L_{nm}$.
       
        \STATE $L_m \leftarrow \emptyset, L_{nm} \leftarrow \emptyset$
        \FOR{img $\in I$}
            \FOR{object $o \in$ img}
                    \STATE $OC \leftarrow$ crop $B_o$ from img.
                    \FOR{$i = 0$ to $S_p$}
                        \STATE $MC \leftarrow$ sample random object $p$ of class $C_o$ and make an image crop.
                        \STATE Append $(OC, MC)$ to $L_m$
                    \ENDFOR
                    
                    \FOR{$i = 0$ to $S_n$}
                        \STATE $NMC \leftarrow$ sample random object $p$ of class $C_p \neq C_o$, and make an image crop.
                        \STATE Append $(OC, NMC)$ to $L_{nm}$
                    \ENDFOR
                    
                    \FOR{$i = 0$ to $S_n$}
                        \STATE $BC \leftarrow$ sample random background patch and make an image crop.
                        \STATE Append $(OC, BC)$ to $L_{nm}$
                    \ENDFOR                    
            \ENDFOR
        \ENDFOR
    \end{algorithmic}        
\end{algorithm}

\subsection{CNN Architecture}

We base our architectural choices on the work of Zagoruyko et al. \cite{zagoruyko2015learning}. This paper introduced CNNs for matching image patches and propose three different architectures for that task: A siamese, pseudo-siamese and a two-channel architecture. We use the two-channel and siamese architectures. In the two-channel architecture, both input images (denoted as $I_A$ and $I_B$) are merged to form a two-channel image, which is the input to our neural network.

The siamese architecture uses two branches that share weights, $I_A$ is input to the left branch, while $I_B$ is input to the right branch. The output of each branch is a feature vector of a fixed size. Both feature vectors from each branch are concatenated to form a single vector that is input to a decision network (formed only of fully connected layers). The idea of a siamese network is that both branches share weights and will learn patch invariant features that are useful for the decision network to produce a matching decision. We use two $96 \times 96$ input images to be matched.

We use the following notation for CNN layers: Conv($N_f$, $F_w \times F_h$) is a convolutional layers with $N_f$ filters of width $F_w$ and height $F_h$. MP($P_w$, $P_h$) is a max-pooling layer with sub-sampling size of $P_w \times P_h$, and FC($n$) is a fully connected layers with $n$ output neurons.

We designed four CNN architectures, two using a 2-Channel approach, and two using a Siamese architecture. We obtained these architectures by performing grid search over a defined set of variations, including depth, number of filters, and filter size. We now describe the 2-Channel architectures:

\begin{itemize}
    \item \textbf{2-Channel CNN Class}. This network is designed to output a binary matching decision (match or non-match), with a two-element output probability distribution given by a softmax function. The full network architecture is Conv($16$, $5 \times 5$)-MP($2$, $2$)-Conv($32$, $5 \times 5$)-MP($2$, $2$)-Conv($32$, $5 \times 5$)-MP($2$, $2$)-Conv($16$, $5 \times 5$)-MP($2$, $2$)-FC($64$)-FC($32$)-FC($2$). The network is trained using a categorical cross-entropy loss function, as the matching decision is formulated as a classification problem.
    \item \textbf{2-Channel CNN Score}. This network outputs a matching score in the range $[0, 1]$ with a sigmoid function. The full network architecture is Conv($16$, $5 \times 5$)-MP($2$, $2$)-Conv($32$, $5 \times 5$)-MP($2$, $2$)-Conv($32$, $5 \times 5$)-MP($2$, $2$)-Conv($16$, $5 \times 5$)-MP($2$, $2$)-FC($64$)-FC($32$)-FC($1$). This network is trained using binary cross-entropy loss function, and the activation at the output is sigmoid.    
\end{itemize}

The Siamese architectures are based on branches with configurations Conv($16$, $5 \times 5$)-MP($2$, $2$)-Conv($32$, $5 \times 5$)-MP($2$, $2$)-Conv($64$, $5 \times 5$)-MP($2$, $2$)-Conv($32$, $5 \times 5$)-MP($2$, $2$)-FC($96$)-FC($96$). The output feature vector contains 96 elements, and the output activation is sigmoid. From this branch architecture we derive the following architectures:

\begin{itemize}
	\item \textbf{Siamese CNN Class}. Both branch outputs are concatenated to form a 192 element vector, that is passed through a decision network with configuration FC($64$)-FC($2$). The output activation in this case is softmax. This network is trained with a categorical cross-entropy loss.
	\item \textbf{Siamese CNN Score}. Same as the previous architecture, but the decision network has configuration FC($64$)-FC($1$), with a sigmoid output activation. This network is trained with a binary cross-entropy loss.
\end{itemize}

All four architectures were obtained by doing grid search over a varying number of layers, convolutional filters and fully connected neurons. The categorical and binary cross-entropy loss functions are the same for the case of binary classification, with the only difference being the application to a single output or to a two-element vector (see Eq \ref{ceLoss}).

\begin{equation}
    L(y, \hat{y}) = - \sum_i y_i \log(\hat{y}_i) = -y_0 \log(\hat{y}_0) - (1 - y_0) \log(1 - \hat{y}_0)
    \label{ceLoss}
\end{equation}

All architectures use ReLU activations, except at the output layers, and are trained on the same dataset, and learn to discriminate sonar image patches. Dropout \cite{srivastava2014dropout} is used after every fully connected layer (except at outputs). We also evaluated the use of Batch Normalization \cite{ioffe2015batch} but Dropout is superior in achieving good generalization performance. Our original design was the class output networks, posed as a classification problem, which works well but interpretation of the output is not trivial as the scores are correlated with the classification outputs. This motivated us to explore the scoring architecture that outputs a score directly that can be separated to obtain a matching decision by a simple threshold . Note that \cite{zagoruyko2015learning} uses networks that only output a score, while we both evaluate continuous score and discrete classification outputs.

\subsection{Training}

Our networks are trained using stochastic gradient descent from random initialized weights. We train using mini-batch gradient descent using a batch size of $b = 128$ images. We adopt the ADAM optimizer \cite{kingma2014adam} for accelerated training and learning rate decay. The initial learning rate is $\alpha = 0.1$. All networks are trained for 5 epochs. We tuned this value on a validation set with early stopping.

In order to prevent the network from learning patterns in the order images are presented, we augment the dataset and for each training pair $(A, B)$, we add the pair $(B, A)$ to the augmented training set. No other data augmentation was used.

\section{Experimental Evaluation}

\subsection{Data}

We have captured a dataset of marine debris objects in our water tank with an ARIS Explorer 3000 (FLS). This dataset consists of 2072 images with $\sim 2500$ total object instances labeled in 9 classes (Metal Cans, Bottles, Drink Carton, Metal Chain, Propeller, Tire, Hook, Valve, Background). On this dataset we ran our matching pair generation algorithm (Algorithm \ref{pairGenerationAlgorithm}). In order to evaluate generalization performance of our networks, we split the dataset according to object class, before obtaining train and test splits. This generated two datasets:

\begin{itemize}
	\item \textbf{Dataset D}: In this dataset the train and testing sets are generated with different objects. Classes 0-5 are used to generate the training set, while classes 6-9 are used to generate the test set. All of our matching networks are trained on this training set.
	\item \textbf{Dataset S}: This dataset is generated using all classes, and split into a training and testing set. From this split we only use the testing set to evaluate performance.
\end{itemize}

The training set (from Dataset D) contains 39840 matching and non-matching pairs ($50 \%$ each). Both testing sets (D and S) contain 7440 matching and non-matching pairs, also balanced.  All reported metrics are evaluated on the test set.

\subsection{Matching Performance}

In this section we evaluate raw matching performance. We plot the ROC curve, and report the Area Under the Curve (AUC). We also obtained accuracy scores for the test set. Accuracy for the matching networks was obtained by considering the raw match probability $p$ (second element of the softmax output, or the sigmoid output score) and taking the class with maximum probability (Eq \ref{eqArgmax}).

\begin{equation}
    c = \argmax \{ 1 - p, p \}
	\label{eqArgmax}
\end{equation}

We compare both our matching networks with the state of the art keypoint detectors and feature extractors, namely SIFT \cite{lowe2004distinctive}, SURF \cite{bay2006surf}, ORB \cite{rublee2011orb} and AKAZE \cite{alcantarilla2011fast}. SIFT and SURF represent the best keypoint detectors for optical images, while ORB was chosen to evaluate its binary features. While it is known \cite{hurtos2012fourier} that these algorithms do not perform well in sonar, there is no other comparison point, as no keypoint detectors have been developed specifically for sonar images. We also compare with Machine Learning (ML) based methods, namely a Support Vector Machine (SVM) as classifier, a Support Vector Regressor (SVR) to regress a score, and Random Forest (RF) classifier and regressor.

Accuracy for keypoint algorithms is obtained by considering a positive match when the ratio test \cite{lowe2004distinctive} gives at least 1 good match. If there are no good matches, then we output a negative match. This threshold is low on purpose to evaluate the best performance of a keypoint matching system.

As our dataset is generated using one type of matching pairs and two different types of non-matching pairs, we also compute the accuracy over each kind of match. This is reported as "Obj-Obj $+$" for Object-Object matching pairs, "Obj-Obj $-$" for Object-Object non-matching pairs, and "Obj-Bg $-$" for Object-Background non-matching pairs.

Table \ref{classicMatchersComparison} displays the main results, only considering the best performing matching networks. Our methods have considerably higher AUC and mean accuracy, which shows that using neural networks for matching sonar images does have a considerable improvement over the state of the art. The 2-Chan CNN Class network has an advantage of $22.4$ AUC percentage points over ORB, with a corresponding increase of $31.3$ accuracy percentage points.

Classical keypoint detectors match sonar images with a chance that is slightly better than chance, with the best classical method being ORB with $0.682$ AUC. Both our matching CNNs outperform classic matches by a considerable margin. Our class matcher also outperforms the scoring matcher. This is due to the fact that scoring matcher is considerably harder to train because the sigmoid output easily saturates. This also contrasts with the results from \cite{zagoruyko2015learning} as they use $\{ +1, -1\}$ scoring with a hinge loss. Our results show that a softmax output with cross-entropy loss can outperform a saturating non-linearity.

Machine Learning methods also perform poorly, but better than keypoint matching. A RF classifier has the highest non-CNN AUC at $0.795$, but their Obj-Obj positive accuracy is considerably poor than the alternatives. SVM and SVE have AUC that is close to keypoint matching.

Classic matchers have a higher Obj-Obj positive accuracy, and this can be explained by overconfident predictions that classify too many pairs as positive matches. This can easily produce high accuracy for positive matches, but will hurt the performance of negative matches. Our matching networks seem to be more balanced, but still their lowest accuracy is when they need to predict a positive match. Both results show the difficulty of matching sonar images. ML methods suffer from the opposite, where they are very accurate for negative matches, but suffer in accuracy for positive matches.

Fig. \ref{rocDifferentObjectComparison} shows the corresponding ROC curves for the classic matchers, ML-based methods,  and our best performing networks. The positive class probability $p$ of Class matching networks and SVM/RF-Class is used to construct the ROC curve, while for Score matching networks and SVR/RF-score the raw score is considered to produce the curve. For keypoint detectors we vary the minimum number of good keypoint matches to declare a positive match. Keypoint matchers have a curve that is very close to random chance, while our methods are closer to a perfect matcher. There is still a considerable room for improvement in the sonar image matching problem. All tested methods produce results that are better than random chance, and RF-based methods are superior when compared to keypoint matching, but still our matching networks are considerably better.

\begin{table*}[htb]
	\centering
	
	\begin{tabular}{|c|c|c|c|c|c|}
		\hline 
		Method 	& AUC	 	& Mean Accuracy & Obj-Obj $+$ Acc 	& Obj-Obj $-$ Acc 	& Obj-Bg $-$ Acc\\ 
		\hline 
		SIFT	& $0.610$ 	& $54.0 \%$ & $74.5 \%$ 		&  $43.6 \%$			&  $44.0 \%$ \\
		SURF	& $0.679$	& $48.1 \%$ & $89.9 \%$ 		&  $18.6 \%$			&  $35.9 \%$ \\
		ORB		& $0.682$	& $54.9 \%$ & $72.3 \%$ 		&  $41.9 \%$			&  $60.5 \%$ \\
		AKAZE	& $0.634$	& $52.2 \%$ & $95.1 \%$ 		&  $4.8 \%$				&  $56.8 \%$ \\
		\hline
		\hline
		RF-Score	& $0.741$	& $57.6 \%$ & $22.5 \%$ 		&  $88.2 \%$				&  $97.2 \%$ \\
		RF-Class	& $0.795$	& $69.9 \%$ & $12.5 \%$ 		&  $97.7 \%$				&  $99.7 \%$ \\
		\hline
		\hline
		SVR-Score	& $0.663$	& $70.5 \%$ & $57.2 \%$ 		&  $66.6 \%$				&  $87.5 \%$ \\
		SVM-Class	& $0.652$	& $67.1 \%$ & $54.4 \%$ 		&  $69.1 \%$				&  $90.5 \%$ \\
		\hline
		\hline
		2-Chan CNN Class & $0.910$	& $86.2 \%$ & $67.3 \%$ &  $95.2 \%$	&  $96.1 \%$ \\
		2-Chan CNN Score & $0.894$	& $82.9 \%$ & $68.0 \%$ &  $96.1 \%$		&  $84.5 \%$ \\
		\hline
	\end{tabular}
	\caption{Comparison of classic keypoint algorithms for matching versus our two best performing matching networks. Area Under the ROC Curve (AUC), Accuracy at match threshold zero, and Accuracy for each match type is reported for Test Set \textbf{D}.}
	\label{classicMatchersComparison}
\end{table*}

Tables \ref{sameObjectComparison} and \ref{differentObjectComparison} show a comparison of all our matching networks on the \textbf{S} and \textbf{D} datasets. Corresponding ROC curves are shown in Fig. \ref{2ChanCNNROC} and Fig. \ref{SiameseCNNROC}. 

Looking at Fig. \ref{2ChanCNNROC}, we can see that network 2-Chan CNN Class performs the best when compared to the scoring network, but this trend reverses when looking at Siamese networks (Fig. \ref{SiameseCNNROC}), as the highest AUC is obtained by Siamese CNN Score. When comparing performance on \textbf{S} and \textbf{D} datasets, we show that performance slightly increases when evaluating on the same objects as the training set (Test set \textbf{S}). This is expected and shows a slight amount of overfit to the training set objects. But generalization performance to unseen objects (Test set \textbf{D}) is still good.

\begin{table}[htb]
	\centering
	
	\begin{tabular}{|c|c|c|c|c|}
		\hline 
		Network Type 	& Output 	& Test Objects 		& AUC	 		& Mean Accuracy \\ 
		\hline 
		2-Chan CNN		& 2 Class 		& Different		& $0.910$ 		& $86.2\%$ \\
		2-Chan CNN		& Score 		& Different		& $0.894$		& $82.9\%$\\
		\hline	
		\hline 
		Siamese CNN		& 2 Class 		& Different		& $0.855$	 	& $82.9 \%$ \\
		Siamese CNN		& Score 		& Different		& $0.826$	 	& $77.0 \%$\\		
		\hline
	\end{tabular}
	\caption{Accuracy and Area Under the ROC Curve (AUC) metrics for Test Set \textbf{D}}
	\label{differentObjectComparison}
\end{table}

\begin{table}[htb]
	\centering
	
	\begin{tabular}{|c|c|c|c|c|}
		\hline 
		Network Type 	& Output 	& Test Objects & AUC	 		& Mean Accuracy \\ 
		\hline 	
		2-Chan CNN		& 2 Class 		& Same			& $0.944$	 	& $ 86.7\%$ \\
		2-Chan CNN		& Score 		& Same			& $0.934$	 	& $ 85.4\%$\\		
		\hline
		\hline
		Siamese CNN		& 2 Class 		& Same			& $0.864$ 		& $ 75.8\%$ \\
		Siamese CNN		& Score 		& Same			& $0.895$		& $ 80.6\%$\\	
		\hline 
	\end{tabular}
	\caption{Accuracy and Area Under the ROC Curve (AUC) metrics for Test Set \textbf{S}}
	\label{sameObjectComparison}
\end{table}

\pgfplotsset{cycle list/Set1-6}

\pgfplotscreateplotcyclelist{matiasList}{%
	{color=red,solid},
	{color=blue,solid},
	{color=green,solid},
    {color=black,solid},
	{color=red,densely dashed},
	{color=blue,densely dashed},
	{color=green,densely dashed},
	{color=black,densely dashed},
	{color=red,densely dotted},
	{color=blue,densely dotted},
	{color=green,densely dotted},
	{color=black,densely dotted}}

\begin{figure}[htb]
	\begin{tikzpicture}
	\begin{axis}[height = 0.24\textheight, width = 0.48\textwidth, xlabel={False Positive Rate}, ylabel={True Positive Rate}, xmin=0, xmax=1.0, ymin=0, ymax=1.0, ymajorgrids=true, xmajorgrids=true, grid style=dashed, legend pos = outer north east, legend style={font=\scriptsize, at = {(0.1, 1.5)}}, cycle multiindex* list={matiasList}, legend columns = 2]
	\addplot +[no markers, raw gnuplot] gnuplot {
		plot 'newData/MatchingCNNDropoutClassNoAugment-DiffObjects-ROCCurve.csv' using 2:3 smooth sbezier;
	};
	\addlegendentry{CNN 2-Chan Class}
	
	\addplot +[no markers, raw gnuplot] gnuplot {
		plot 'newData/MatchingCNNDropoutScoreBCENoAugment-DiffObjects-ROCCurve.csv' using 2:3 smooth sbezier;
	};
	\addlegendentry{CNN 2-Chan Score}
	
	\addplot +[no markers, raw gnuplot] gnuplot {
		plot 'mlData/new/svmProbsROCCurve.csv' using 2:3 smooth sbezier;
	};
	\addlegendentry{SVM-Class}
	
	\addplot +[no markers, raw gnuplot] gnuplot {
		plot 'mlData/new/svrROCCurve.csv' using 2:3 smooth sbezier;
	};
	\addlegendentry{SVR-Score}
	
	\addplot +[no markers, raw gnuplot] gnuplot {
		plot 'mlData/new/randomForestClassifierROCCurve.csv' using 2:3 smooth sbezier;
	};
	\addlegendentry{RF-Class}
	
	\addplot +[no markers, raw gnuplot] gnuplot {
		plot 'mlData/new/randomForestROCCurve.csv' using 2:3 smooth sbezier;
	};
	\addlegendentry{RF-Score}
	
	\addplot+[mark = none] table[x  = fpr, y  = tpr, col sep = space] {newData/siftMatcherPRCurve.csv};
	\addlegendentry{SIFT}
	\addplot+[mark = none] table[x  = fpr, y  = tpr, col sep = space] {newData/surfMatcherPRCurve.csv};
	\addlegendentry{SURF}
	\addplot+[mark = none] table[x  = fpr, y  = tpr, col sep = space] {newData/orbMatcherPRCurve.csv};
	\addlegendentry{ORB}	
	\addplot+[mark = none] table[x  = fpr, y  = tpr, col sep = space] {newData/akazeMatcherPRCurve.csv};
	\addlegendentry{AKAZE}
	\draw[gray] (axis cs:0,0) -- (axis cs:1,1);
	\end{axis}        
5	\end{tikzpicture}
	\caption{ROC curve comparing our matching networks with ML-based methods (SVM/SVR and Random Forest, RF) and keypoint matching methods (SIFT, SURF, ORB and AKAZE). Our methods clearly outperform other methods when used to match sonar images. These curves were evaluated on the \textbf{D} dataset. The grey line represents random chance limit.}
	\label{rocDifferentObjectComparison}
\end{figure}
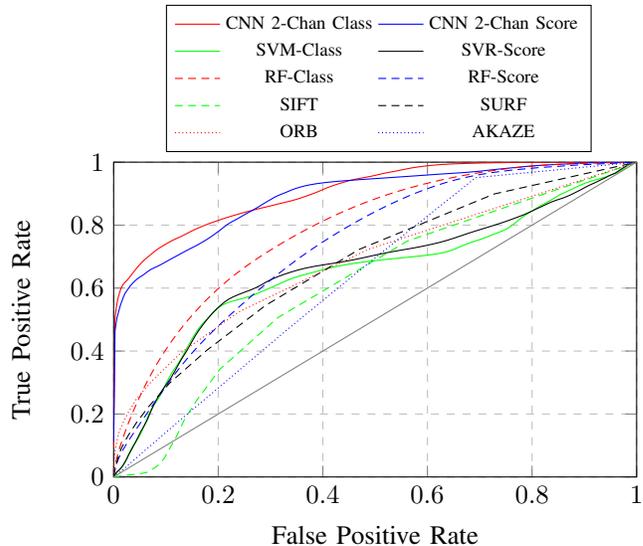

\begin{figure*}[htb]
	\begin{subfigure}[b]{0.49 \textwidth}
	\begin{tikzpicture}
	\begin{axis}[height = 0.18\textheight, width = \textwidth, xlabel={False Positive Rate}, ylabel={True Positive Rate}, xmin=0, xmax=1.0, ymin = 0.4, ymax = 1.0, ytick = {0.4, 0.5, 0.6, 0.7, 0.8, 0.85, 0.9, 0.95, 1.0}, ymajorgrids=true, xmajorgrids=true, grid style=dashed, legend pos = south east, legend style={font=\scriptsize}, cycle list name=Set1-6]        
	
	\addplot +[no markers, raw gnuplot] gnuplot {
		plot 'newData/MatchingCNNDropoutClassNoAugment-SameObjects-ROCCurve.csv' using 2:3 smooth sbezier;
	};
	\addlegendentry{2-Chan CNN Class \textbf{S} (AUC 0.94)}
	
	\addplot +[no markers, raw gnuplot] gnuplot {
		plot 'newData/MatchingCNNDropoutScoreBCENoAugment-SameObjects-ROCCurve.csv' using 2:3 smooth sbezier;
	};
	\addlegendentry{2-Chan CNN Score \textbf{S} (AUC 0.93)}
	
	\addplot +[no markers, raw gnuplot] gnuplot {
		plot 'newData/MatchingCNNDropoutClassNoAugment-DiffObjects-ROCCurve.csv' using 2:3 smooth sbezier;
	};
	\addlegendentry{2-Chan CNN Class \textbf{D} (AUC 0.91)}
	
	\addplot +[no markers, raw gnuplot] gnuplot {
		plot 'newData/MatchingCNNDropoutScoreBCENoAugment-DiffObjects-ROCCurve.csv' using 2:3 smooth sbezier;
	};
	\addlegendentry{2-Chan CNN Score \textbf{D} (AUC 0.89)}
	
	\end{axis}        
	\end{tikzpicture}
	\caption{2-Channel CNN}
	\label{2ChanCNNROC}
	\end{subfigure}
	\begin{subfigure}[b]{0.49 \textwidth}
	\begin{tikzpicture}
	\begin{axis}[height = 0.18\textheight, width = \textwidth, xlabel={False Positive Rate}, ylabel={True Positive Rate}, xmin=0, xmax=1.0, ymin = 0.4, ymax = 1.0, ytick = {0.4, 0.5, 0.6, 0.7, 0.8, 0.85, 0.9, 0.95, 1.0}, ymajorgrids=true, xmajorgrids=true, grid style=dashed, legend pos = south east, legend style={font=\scriptsize}, cycle list name=Set1-6]        
	
	\addplot +[no markers, raw gnuplot] gnuplot {
		plot 'newData/MatchingSiameseCNNDenseClass-DiffObjects-ROCCurve.csv' using 2:3 smooth sbezier;
	};
	\addlegendentry{Siamese CNN Class \textbf{S} (AUC 0.86)}
	
	\addplot +[no markers, raw gnuplot] gnuplot {
		plot 'newData/MatchingSiameseCNNDenseScore-SameObjects-ROCCurve.csv' using 2:3 smooth sbezier;
	};
	\addlegendentry{Siamese CNN Score \textbf{S} (AUC 0.89)}
	
	\addplot +[no markers, raw gnuplot] gnuplot {
		plot 'newData/MatchingSiameseCNNDenseClass-SameObjects-ROCCurve.csv' using 2:3 smooth sbezier;
	};
	\addlegendentry{Siamese CNN Class \textbf{D} (AUC 0.85)}
	
	\addplot +[no markers, raw gnuplot] gnuplot {
		plot 'newData/MatchingSiameseCNNDenseScore-DiffObjects-ROCCurve.csv' using 2:3 smooth sbezier;
	};
	\addlegendentry{Siamese CNN Score \textbf{D} (AUC 0.82)}
	
	\end{axis}        
	\end{tikzpicture}
	\caption{Siamese CNN}
	\label{SiameseCNNROC}
	\end{subfigure}

	\caption{ROC curves comparing 2 Channel and Siamese matching networks on test sets \textbf{S} and \textbf{D}. This comparison shows the difference in performance Test set \textbf{S} was generated with the same objects as in the training set. Test set \textbf{D} was generated with different objects than in the training set.}
	
	\label{rocComparisonAcrossObjects}
\end{figure*}
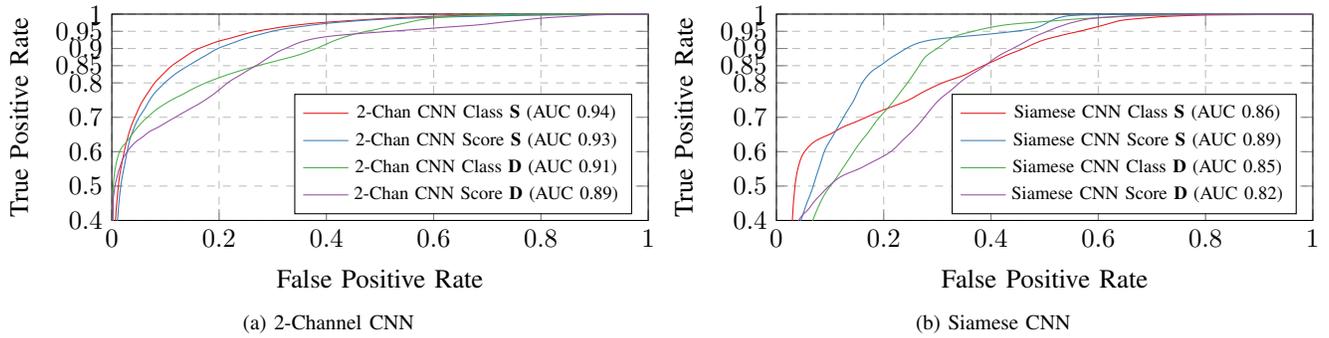

Finally, Table \ref{accuracyByMatchType} offers a breakdown of accuracy in our matching networks over the three different match pairs on both datasets \textbf{S} and \textbf{D}. When evaluated in different objects, all networks have decreased performance on Object to Object positive matches, while at the same time having adequate performance in both negative cases. For evaluation in the same objects as the training set, this trend reverses (as expected) and the largest accuracies are reported in the Object to Object positive matches. This indicates that the networks are slightly overfitting the objects in the training set, but still they do provide an improvement over keypoint matching and ML-based methods in unseen objects. Matching an object to a different view of the same object is also a hard problem, and discarding matches from different objects or to background is easier.

\begin{table*}[htb]
    \centering
    
    \begin{tabular}{|c|c|c|c|c|c|c|c|}
        \hline 
         & & \multicolumn{3}{|c|}{Different Objects} & \multicolumn{3}{|c|}{Same Objects} \\
         \hline
        Network Type	&  Output 	& Obj-Obj $+$ Acc 	& Obj-Obj $-$ Acc 	& Obj-Bg $-$ Acc & Obj-Obj $+$ Acc 	& Obj-Obj $-$ Acc 	& Obj-Bg $-$ Acc\\ 
        \hline 
        2-Chan CNN		&  2 Class 	& $67.3 \%$ 		&  $95.2 \%$			&  $96.1 \%$ & $86.6 \%$ 		&  $75.7 \%$			&  $97.8 \%$ \\
        2-Chan CNN		&  Score	& $68.0 \%$ 		&  $96.1 \%$			&  $84.5 \%$ & $85.0 \%$ 		&  $77.5 \%$			&  $93.7 \%$  \\
        \hline
        \hline
        Siamese CNN		&  2 Class	& $62.9 \%$ 		&  $89.9 \%$			&  $96.0 \%$ & $92.2 \%$ 		&  $39.4 \%$			&  $95.8 \%$\\
        Siamese CNN		&  Score	& $49.2 \%$ 		&  $84.7 \%$			&  $97.0 \%$ & $89.1 \%$ 		&  $55.3 \%$			&  $97.3 \%$\\
        \hline 
    \end{tabular}
    \caption{Accuracy by matching type, for the \textbf{S} and \textbf{D} datasets}
    \label{accuracyByMatchType}
\end{table*}

\subsection{Discussion}

Our 2-Channel CNN Class matching networks performs well, with an AUC of 0.91 when evaluated on unseen data, and AUC of 0.94 on a dataset that shares objects with the train set. We have not seen any previous work claiming to match sonars images with such performance. Classic methods (SIFT/SURF and ORB) are known to work poorly in sonar, but still they are being used for registration in a large part of the literature \cite{kim2005mosaicing} \cite{negahdaripour2011dynamic} \cite{vandrish2011side}. We have not seen quantitative results of keypoint matching in sonar images, and our extensive evaluation is useful for comparison.

One clear limitation of our evaluation is the small size of our datasets (39K for training and 7K for testing). We believe our networks could greatly benefit from a bigger dataset, containing more variability in objects as well as more object views. A dataset captured in real underwater conditions is ideal but hard to produce.


\section{Conclusions}

We have proposed the use of CNNs to learn a matching function for sonar images. Our results show that our 2-Chan CNN Class network can match FLS image patches with high accuracy (AUC 0.91), while classic keypoint matching methods can do it only with low accuracy, slightly better than random chance (AUC 0.61 to 0.68). ML-based methods (SVM and Random Forests) are also inferior (AUC 0.65 to 0.80).


We believe that our results are appropriate for the small (39K samples) training dataset that we possess, and they could improve if more data and object/background variability is available. Our dataset was captured under laboratory conditions in a small water tank containing household garbage objects, and our work can surely be improved with a dataset captured in real underwater scenes.

Our method is limited by available number of images and objects in them. Our network architecture could fail to generalize with other objects, specially if their shape is radically different. Background is also a concern, as the background in our water tank is not representative of a typical underwater environment.

We expect that AUV perception will benefit from our work and open possibilities of advanced CNN-based matching and registration methods. Our work can be applied to any kind of sonar, as we did not make assumptions on the kind of image produced by the sonar.

As future work, we would like to build a similarity function for sonar images instead of making a binary decision and learn from image patches in a unsupervised way. The best case is to learn from a dataset that contains automatically matched scenes with another sensor (like depth in optical images) as \cite{zbontar2016stereo} and \cite{zagoruyko2015learning} do.

\section*{Acknowledgments}

This work has been partially supported by the FP7-PEOPLE-2013-ITN project ROBOCADEMY (Ref 608096) funded by the European Commission. The author would like to thank Leonard McLean for his help in capturing data used in this paper.

\bibliographystyle{IEEEtran}
\bibliography{biblio}

\end{document}